\documentclass[journal]{IEEEtran}
\usepackage{multirow} 
\usepackage{booktabs}
\usepackage{makecell}
\usepackage{cite}
\usepackage{amsmath} 
\usepackage[ruled]{algorithm2e} 
\usepackage{array}
\usepackage{graphicx}
\usepackage{float}
\usepackage{amsthm,amsmath,amssymb}
\usepackage[table,xcdraw]{xcolor}
\graphicspath{{./figures/}}
\usepackage{url}  

\ifCLASSOPTIONcompsoc
  \usepackage[caption=false,font=normalsize,labelfont=sf,textfont=sf]{subfig}
\else
  \usepackage[caption=false,font=footnotesize]{subfig}
\fi

\ifCLASSINFOpdf
\else
\fi

\begin{document}

\title{Seismic First Break Picking in a Higher Dimension Using Deep Graph Learning}

\author{Hongtao~Wang, Li~Long, Jiangshe~Zhang, Xiaoli~Wei, Chunxia~Zhang, Zhenbo~Guo.
\thanks{Corresponding authors: Jiangshe Zhang}
\thanks{H.T. Wang, J.S. Zhang, X.L. Wei, L. Long, C.X. Zhang are with the School of Mathematics and Statistics, Xi'an Jiaotong University, Xi'an, Shaanxi, 710049, P.R.China.}
\thanks{Z.B. Guo is with China National Petroleum Corp Bureau of Geophysical Prospecting Inc, Geophysical Technology Research Center Zhuozhou, Hebei, 072750, P.R.China.}
\thanks{The research is supported by the National Natural Science Foundation of China under grant 12371512.}}

%
%

\markboth{Journal of \LaTeX\ Class Files,~Vol.~14, No.~8, August~2015}%
{Wang \MakeLowercase{\textit{et al.}}: GNN for FB Picking}

\maketitle
\begin{abstract}
Contemporary automatic first break (FB) picking methods typically analyze 1D signals, 2D source gathers, or 3D source-receiver gathers. Utilizing higher-dimensional data, such as 2D or 3D, incorporates global features, improving the stability of local picking. Despite the benefits, high-dimensional data requires structured input and increases computational demands. Addressing this, we propose a novel approach using deep graph learning called DGL-FB, constructing a large graph to efficiently extract information. In this graph, each seismic trace is represented as a node, connected by edges that reflect similarities. To manage the size of the graph, we develop a subgraph sampling technique to streamline model training and inference. Our proposed framework, DGL-FB, leverages deep graph learning for FB picking. It encodes subgraphs into global features using a deep graph encoder. Subsequently, the encoded global features are combined with local node signals and fed into a ResUNet-based 1D segmentation network for FB detection. Field survey evaluations of DGL-FB show superior accuracy and stability compared to a 2D U-Net-based benchmark method.
\end{abstract}

\begin{IEEEkeywords}
First break picking, Seismic data processing, Graph neural network
\end{IEEEkeywords}

\IEEEpeerreviewmaketitle

\section{Introduction}
\IEEEPARstart{F}{irst} break (FB) picking, pivotal for statics corrections in seismic data processing \cite{yilmaz2001seismic}, has seen various automatic methods like STA/LTA \cite{allen1978automatic}, CNN \cite{hollander2018using}, and U-Net architectures \cite{hu2019first, han2021first}. These techniques, analyzing seismic data samples across 1D, 2D, and 3D formats, leverage data arrangement to enhance analytic similarity. For instance, U-Net capitalizes on the continuous texture of first arrivals in 2D images for efficient picking. However, limiting data to 2D or 3D underutilizes the potential of computers for higher-dimensional observation. 

To model seismic data in higher dimensions, we introduce a novel modeling approach to analyze seismic data on a graph. Each seismic trace is treated as a node, and edges are established between nodes based on the near midpoints of corresponding sources and receivers. From a graph perspective, a 2D shot gather is identified as a specific instance of graph sampling, where signals originating from the same source point and receiver lines constitute a subgraph. Consequently, graph analysis provides a more generalized methodology. While graph neural networks have been applied in geophysics for tasks like seismic event classification \cite{kim2021graph} and attribute regression \cite{bloemheuvel2023graph}, these models often oversimplify graph construction, limiting applicability for FB picking and neglecting global information integration.

Consequently, graph analysis provides a more generalized methodology. The application of graph neural network theory in geophysics has gained traction. Researchers have endeavored to construct graphs using the coordinates of source and receiver points for addressing tasks such as seismic event classification \cite{kim2021graph} and seismic attribute regression \cite{bloemheuvel2023graph}. In these approaches, traces are considered as nodes, and edges are formed among nodes associated with the same source and adjacent receiver points. Subsequently, graph neural networks are employed to encode interrelated seismic traces, with the encoded features directed to a downstream task decoder. However, the graph construction rules in these methods are overly simplistic, focusing solely on adjacent picking points corresponding to the same shot gather. Consequently, these methods are not directly applicable to the first break picking task. Notably, when compared to 2D image-based first break picking methods, the constructed graph lacks the incorporation of global information to enhance the accuracy of picking individual traces.

To address this issue, we propose a novel picking framework using the deep graph learning technique. Specifically, we first build a huge graph for the whole survey. Second, to boost the training and inference processes, the subgraphs are sampled from the huge graph. Third, a graph neural network is built to encode the global information of the subgraph. Finally, the combined information of global information and local information is fed to a ResUNet to obtain the FB. To estimate the performance, we compared our automatic picking results with manual picking results and a 2D benchmark method on a field dataset.

\begin{figure*}[!ht]
  \centering
  \includegraphics[width=\textwidth]{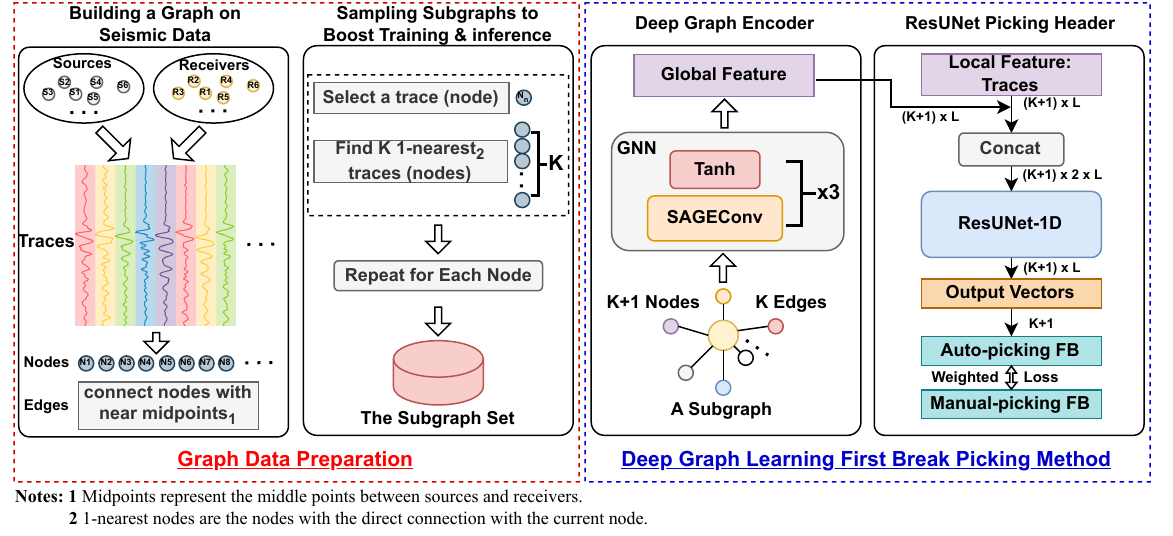}
  \caption{A showcase of the FB picking task. The left subfigure shows the FB of a single trace. The right subfigure indicates a high correlation among the FBs of the adjacent traces. The red dotted box indicates the location of the single-trace signal in the left subfigure.}
  \label{fig: mainflow}
\end{figure*}

\section{Method}
To tackle the challenge of picking the first break in a higher dimension, we propose a deep graph learning-based automatic picking method called DGL-FB. This method incorporates two techniques: a deep graph convolution encoder and the ResUNet picking header. Specifically, a huge graph is generated from survey data, with each node representing a trace and the edges connecting traces with strong correlations. Then, we sample a series of subgraphs as the input for DGL-FB. In DGL-FB, the subgraph is initially encoded into a global feature and then concatenated with the signal of the central node. This concatenated feature is then fed into the ResUNet to produce the final first break picking of the central node. 

\subsection{Graph Data Preparation}
In mathematics and computer science, an undirected graph $(G)$ is a collection of nodes $(V)$ and edges $(E)$ that connect pairs of nodes: $G=(V,E)$.
In the seismic data, we assume that each trace is a node, and the edges connect the pairs of the nodes with high correlations. We establish connections between the traces with nearby midpoints \cite{diebold1981traveltime}. The $i$th trace connects with the traces whose midpoints are the k-nearest neighbors of the midpoint of the $i$th trace. 
Since the trace number is the million level, the built graph is too huge and cannot be fed into the memory of the computer to process. Thus, we have to sample a few subgraphs as the input. In this study, we sample a node and its $K$ 1-nearest neighbor nodes in the huge graph as a sample point (or a subgraph). Therefore, there is $K+1$ nodes and $K$ edges in the sampled subgraph. We repeat this opearation for each node and obtain a subgraph set. 

\subsection{DGL-FB} 
To process the graph data, we utilize deep graph learning and propose a novel first break picking framework named DGL-FB. There are two steps in DGL-FB as shown in Figure~\ref{fig: mainflow}. First, a subgraph is encoded as global features. Second, DGL-FB combines the global features with the local features (trace signals of the subgraph nodes) and segments the combined features to first break times. 

\begin{figure}[!htb]
  \centering
  \includegraphics[width=0.45\textwidth]{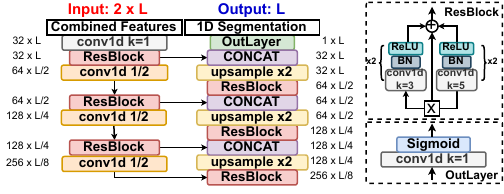}
  \caption{ResUNet-1D Network Structure}
  \label{fig: resunet}
\end{figure}

In the deep graph encoder, we adapt GraphSAGE (SAmple and aggreGatE) method \cite{hamilton2017inductive}. Concretely, we choose SAGEConv layer with LSTM aggregator as the basic layer. We denote $\mathbf{h}^{l}_{v}$ as the representation of node $v$ at the $l$th layer and define $\mathbf{h}^{(0)}_{v}$ as the trace signal corresponding to the node $v$. There are two steps in a SAGEConv layer. First, the information of adjacent nodes is aggregated using LSTM layer \cite{hochreiter1997long}.

\begin{equation}
  \mathbf{h}^{(l+1)}_{\mathcal{N}(v)} \leftarrow \text{LSTM}_l(\{w_{uv}\mathbf{h}^{(l)}_u, \forall u \in \mathcal{N}(v)\}),
  \label{Eq: LSTM_agg}
\end{equation}
where $\mathcal{N}(v)$ represents the set of adjacent nodes of node $v$, and $w_{uv}$ denotes the weight of the edge between the node $u$ and the node $v$. In this study, we compute the weights based the midpoint distances:
\begin{equation}
  w_{uv} = \frac{3}{4} \times \left(1-\frac{d_{uv}^2}{max_{u}\{d_{uv}\}^2}\right),
  \label{Eq: weights}
\end{equation}
where we adapt Epanechnikov quadratic kernel method, and $d_{uv}$ is the distance between the midpoints of the node $u$ and the node $v$.
Second, the aggregated features of adjacent nodes are combined with the current node $v$, and fed to a full-connection layer:
\begin{equation}
  \mathbf{h}^{(l+1)}_{v} \leftarrow \sigma\left(\mathbf{W}^{l+1}\cdot \text{CONCAT}(\mathbf{h}^{(l)}_{v}, \mathbf{h}^{(l+1)}_{\mathcal{N}(v)})\right).
  \label{Eq: concat_out}
\end{equation}

\begin{figure*}[!htb]
  \centering
  \includegraphics[width=\textwidth]{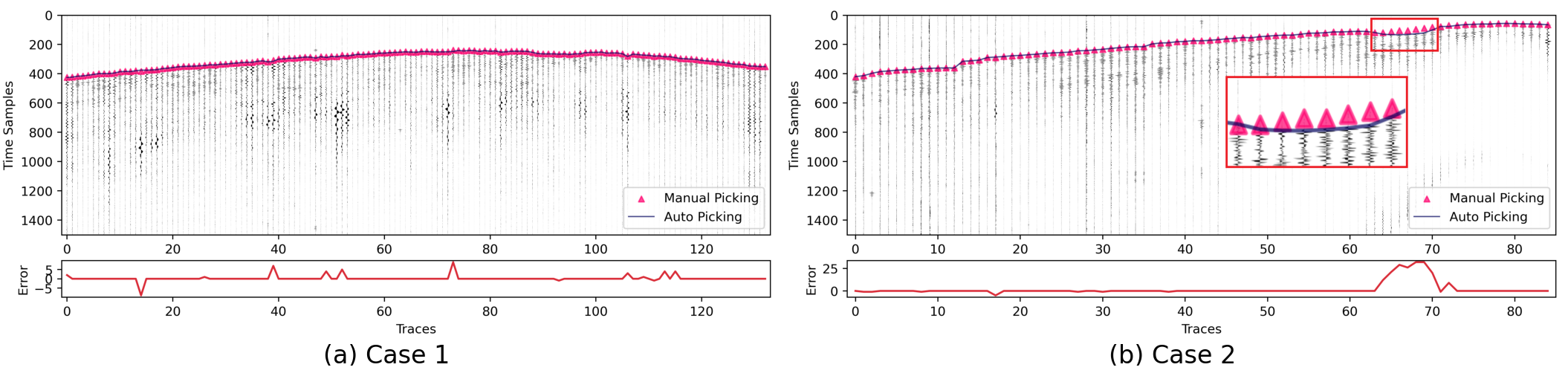}
  \caption{Two picking cases of a common shot gather from the test survey Lalor using DGL-FB.}
  \label{fig: CSG-pick}
  \end{figure*}

Additionally, since the input signals are time series with both positive and negative amplitudes, we adapt TanH activation function to map the feature to range of [-1, 1]:
\begin{equation}
  \text{TanH}(z) = \frac{e^z-e^{-z}}{e^z+e^{-z}}.
  \label{Eq: tanh}
\end{equation}
We repeat the combination of the SAGEConv layer and the TanH activation layer three times to create the deep graph encoder as shown in Figure~\ref{fig: mainflow}. Ultimately, the encoder produces comprehensive features for each node, with the length of the output feature matching the signal of the node.

To integrate global and local features and pick FB, we draw inspiration from the 2D ResUNet method \cite{diakogiannis2020resunet} and introduce the 1D ResUNet for addressing segmentation problems in time series data. The combined features (shape = $2\times L$) are first generated by concatenating the global features and the local features. Then, a 1-D Convolutional layer with kernel size = 1 maps the features to a tensor with channel = 32. Subsequently, there are three combination of ResBlock and the downsampling layer, in which the feature length output for each combination is reduced by half, while the number of channels is doubled. In the encoder of ResUNet-1D, there are three combinations of ResBlocks and upsampling layers, where the feature length output for each combination is doubled, while the number of channels is reduced by half. The upsampled features from the decoder are concatenated with the features output by the resblock in the encoder, based on the channels, to enhance the perception of high-frequency information. Finally, the segmentation of FB (length=L) is output by a output layer. To supervise the output segmentation, we adapt the mask of each trace by assigning a value of 0 to pixels occurring before the FB time, and a value of 1 to the remaining pixels. Additionally, we design a weighted binary cross-entropy (BCE) loss function on a subgraph ($K+1$ nodes), comprising a primary loss (first item) and an auxiliary loss (second item):
\begin{equation}
  L(\mathbf{\hat{y}}, \mathbf{y}) = (1-\lambda) \cdot L_{BCE}(\hat{y}_1, {y}_1) + \lambda \cdot \left(\sum_{k=1}^{K} w_{k}\cdot L_{BCE}(\hat{y}_k, {y}_k) \right),
  \label{Eq: weighted_loss}
\end{equation}
where $\lambda$ is the weight between the main loss and the auxiliary loss (0.5 in our study), and $w_k$ is the weight of the edge between the current predicted trace and the $k$th node used in Eq.~\ref{Eq: LSTM_agg}. 

\section{Experimental Results}
We evaluate DGL-FB on a open-source dataset \cite{pierre2021multi}, which includes four field surveys, named Halfmile, Brunswick, Sudbury, and Lalor, respectively. In this study, Brunswick and Sudbury are the training sets, Halfmile is the validation set, and Lalor is the test set. To maintain consistent input length, we first apply linear moveout (LMO) correction to trim the trace to a length of 128. Then, we normalize the cropped trace to the range of [-1, 1] by dividing it by the absolute value of the maximum. In the training process, we utilize the AdamW optimizator\cite{loshchilov2017decoupled} with decay rate 1e-4, the initial learning rate of 1e-2, and a training minibatch size of 256. We compared our method with the latest 2D U-Net-based benchmark methods\cite{st2023deep}. Compared to the benchmark method with an accuracy of 76.3\%, DGL-FB achieves an accuracy of 81.8\%, showing a 7.2\% improvement. Additionally, root mean square error (RMSE) serves as a good measure of picking stability, and DGL-FB (RMSE=3.24) reduces the RMSE by 99.3\% compared to the benchmark method (RMSE=460.0). We also visualize the picking results of DGL-FB on a common shot gather as shown in Figure~\ref{fig: CSG-pick}. Figure~\ref{fig: CSG-pick} indicates that the DGL-FB method incorporates global information encoded by deep graph encoder, resulting in robust picking results that are nearly identical to manual picking results (Figure~\ref{fig: CSG-pick}a), with some areas even outperforming manual picking results (Figure~\ref{fig: CSG-pick}b).


\section{Conclusion}
This study introduces a novel picking framework called DGL-FB. We evaluate the effectiveness of DGL-FB using field survey datasets and draw the following conclusions.
1) Seismic data can be considered as graph data, and it can be encoded using graph neural network methods to capture the shared global information of subgraphs.
2) The ResUNet-1D technique in DGL-FB effectively integrates global and local features, providing robust FB picking values.
3) The experimental results demonstrate that our method achieves a higher picking accuracy compared to the current 2D benchmark picking method. Furthermore, DGL-FB maintains a higher picking rate while ensuring accuracy.

\section*{Acknowledgment}
The author would like to thank Mr. Pierre-Luc St-Charles from Applied Machine Learning Research Team Mila, Québec AI Institute for providing the open datasets.

\ifCLASSOPTIONcaptionsoff
  \newpage
\fi

\bibliographystyle{IEEEtran}
\bibliography{samplebib}

\end{document}